\documentclass[letterpaper, 10 pt, journal, twoside]{IEEEtran}   % Comment this line out if you need a4paper

\IEEEoverridecommandlockouts                              % This command is only needed if
\usepackage[T1]{fontenc}
\usepackage{graphics} % for pdf, bitmapped graphics files
\usepackage[pdftex]{graphicx} % for postscript graphics files
\usepackage{amsmath} % assumes amsmath package installed
\usepackage{amssymb}  % assumes amsmath package installed
\usepackage{blindtext}
\usepackage{todonotes}
\usepackage{acro}
\usepackage{balance}
\usepackage{siunitx}
\usepackage{subcaption}
\usepackage[noadjust]{cite}
\usepackage{stackengine}
\usepackage{hyperref}
\usepackage[font={footnotesize}]{caption}
\usepackage{bbm}
\usepackage{multirow}
\usepackage{xcolor}
\usepackage[export]{adjustbox}
\providecommand{\DIFdel}[1]{} % Don't show deleted text

\newcommand{\soa}{state-of-the-art }
\newcommand{\etal}{\textit{~et al.}}

\newcommand{\mat}[1]{\boldsymbol{\mathbf{#1}}}
\renewcommand{\u}{\mat{u}}
\newcommand{\y}{\mat{y}}
\newcommand{\g}{\mat{g}}
\newcommand{\params}{\mat\theta}

\DeclareAcronym{il}{short = IL, long = imitation learning}
\DeclareAcronym{rl}{short = RL, long = reinforcement learning}
\DeclareAcronym{cpo}{short = CPO, long = Constrained Policy Optimization}
\DeclareAcronym{trpo}{short = TRPO, long = Trust Region Policy Optimization}
\DeclareAcronym{lrf}{short = LRF, long = laser range finder}
\DeclareAcronym{ril}{short = \mbox{R-IL}, long = reinforced imitation learning}
\DeclareAcronym{gan}{short = GAN, long = General Adversarial Network}
\DeclareAcronym{cnn}{short = CNN, long = convolutional neural network}
\DeclareAcronym{irl}{short = IRL, long = inverse reinforcement learning}
\DeclareAcronym{kl}{short = KL, long = Kullback-Leibler}
\DeclareAcronym{mdp}{short = MDP, long = Markov Decision Process}
\DeclareAcronym{gail}{short = GAIL, long = Generative Adversarial Imitation Learning}

\title{Reinforced Imitation: Sample Efficient Deep Reinforcement Learning for Map-less Navigation by Leveraging Prior Demonstrations}

\author{M. Pfeiffer$^{1*}$, S. Shukla$^{2*}$, M. Turchetta$^{3,4}$, C. Cadena$^{1}$, A. Krause$^{3}$, R. Siegwart$^{1}$, J. Nieto$^{1}$
\thanks{$^*$The authors contributed equally to this work.}
\thanks{The authors are with the $^1$ Autonomous Systems Lab, $^2$ Computer Vision Lab, $^3$ Learning \& Adaptive Systems Group, and $^4$ Max Planck ETH Center for Learning Systems,
ETH Zurich, Zurich, Switzerland. \newline
{\tt \footnotesize{\{pfmark, shuklas, matteotu, cesarc, krausea, rsiegwart, nietoj \}@ethz.ch}}.}
}

\begin{document}

\maketitle

\newlength{\subImageWidth}
\setlength{\skip\footins}{3mm}
\setlength{\abovedisplayskip}{2mm}
\setlength{\belowdisplayskip}{2mm}

%%%%%%%%%%%%%%%%%%%%%%%%%%%%%%%%%%%%%%%%%%%%%%%%%%%%%%%%%%%%%%%%%%%%%%%%%%%%%%%%
\begin{abstract}
This work presents a case study of a learning-based approach for target driven map-less navigation.
The underlying navigation model is an end-to-end neural network which is trained using a combination of expert demonstrations, imitation learning (IL) and reinforcement learning (RL).
While RL and IL suffer from a large sample complexity and the distribution mismatch problem, respectively, we show that leveraging prior expert demonstrations for pre-training can reduce the training time to reach at least the same level of performance compared to plain RL by a factor of 5.
We present a thorough evaluation of different combinations of expert demonstrations, different RL algorithms and reward functions, both in simulation and on a real robotic platform.
Our results show that the final model outperforms both standalone approaches in the amount of successful navigation tasks.
In addition, the RL reward function can be significantly simplified when using pre-training, e.g. by using a sparse reward only.
The learned navigation policy is able to generalize to unseen and real-world environments.
\end{abstract}


%%%%%%%%%%%%%%%%%%%%%%%%%%%% Introduction %%%%%%%%%%%%%%%%%%%%%%%%%%%%%%%%%%%%%%
\section{Introduction}
\label{sec:introduction}

Autonomous navigation in environments where global knowledge of the map is available is nowadays well understood \cite{lavalle2006planning}.
Optimization objectives like, e.g., minimum path length, travel time or safe distance to obstacles can be used to find the optimal path connecting the start and goal position of a robot.
However, full knowledge of the map is not always available in practice, e.g., in  search and rescue applications or rapidly changing environments.
If no reliable environment map can be used for navigation, classical path planning approaches \cite{lavalle2006planning} might fail.
Given only local perception of the robot and a relative target position, robust map-less navigation strategies are required.
In recent years, machine learning techniques --- with neural networks leading the way \cite{pfeiffer2017perception,tai2017virtual,muller2005off} --- have gained importance allowing for the application of end-to-end motion planning approaches.
Instead of splitting the navigation task into multiple sub-modules like, e.g., sensor fusion, obstacle detection, global and local motion planning, end-to-end approaches use a direct mapping from sensor data to robot motion commands which can reduce the complexity during deployment significantly.

Current \soa end-to-end planning approaches can be split in two major groups:
(i) \acf{il} based ones use supervised learning techniques to imitate expert demonstrations as close as possible\footnote{Also known as behavioral cloning}, and (ii) approaches based on \acf{rl} where the agents learn their navigation policy by trial and error exploration combined with reward signals.
\ac{il} is sample efficient and can achieve accurate imitation of the expert demonstrations.
Given the training data, satisfactory navigation models can be found within a few hours of training \cite{pfeiffer2017perception}.
However, it is likely to overfit to the environment and situations presented at training time.
This limits the potential for generalization and the robustness of the policy (distribution mismatch).
\ac{rl} is conceptually more robust --- also in unseen scenarios --- as the agent learns from its own mistakes during training \cite{tai2017virtual}.
The disadvantage of \ac{rl} is its sample inefficiency and missing safety during training, limiting the current utilization to applications where training can be conducted using extremely fast simulators \cite{mnih2016asynchronous}.
As for \ac{rl} training, episodes need to be forward simulated (on- or off-policy), training iterations are significantly more time consuming than in \ac{il}, which reduces the number of training iterations in a given time.
However, \ac{rl} allows to encode desired behavior --- such as reaching the target and avoiding collisions --- specifically in a reward function and does not only rely on suitable expert demonstrations.
In addition, \ac{rl} maximizes the overall expected return on a full trajectory, while \ac{il} treats every observation independently \cite{kuefler2017imitating}, which conceptually makes \ac{rl} superior to \ac{il}.

\begin{figure}[t]
\centering
    \includegraphics[width=\linewidth,trim=0 10 0 0]{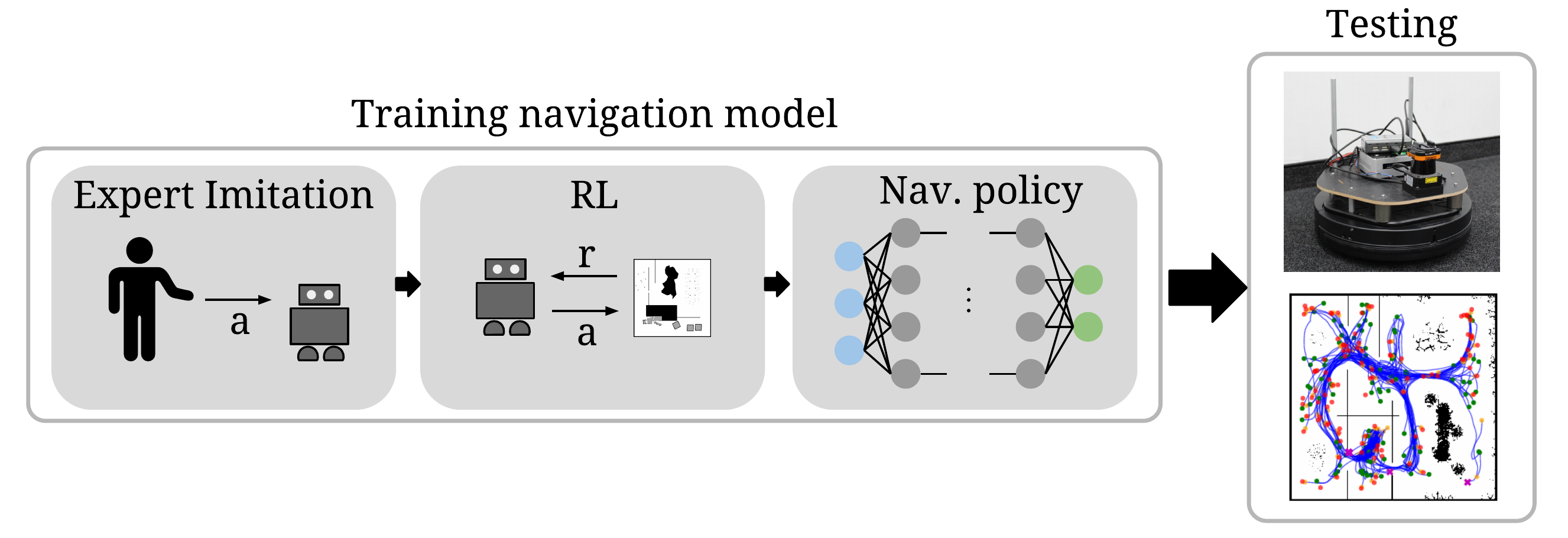}
    \caption{An end-to-end navigation policy is learned from a combination of imitation and reinforcement learning.
    The resulting policy is tested thoroughly in simulation and on a real robotic platform.}
    \label{fig:teaser}
\vspace{-8mm}
\end{figure}

In this work, we present and analyze an approach that combines the advantages of both \ac{il} and \ac{rl}.
It is inspired by human learning, which typically combines the observation of other people and self-exploration \cite{grimes2009learning}.
Our approach, in the following called \textit{\ac{ril}}, combines supervised \ac{il} based on expert demonstrations to pre-train the navigation policy with subsequent \ac{rl}.
For \ac{rl}, we use \ac{cpo} \cite{achiam2017cpo} due to its ability to incorporate constraints during training.
This allows for safer training and navigation, which is especially important for real-world mobile robotics.

We hypothesize that the combination of the two learning approaches yields a more robust policy than pure \ac{il}, and that it is also easier and faster to train than pure \ac{rl}.
In addition, by enforcing the collision avoidance by constraint instead of a fixed penalty in the reward function, the amount of collisions during training and testing should be decreased.
To the best of our knowledge, this is the first work to explore this combination for robot navigation and also to apply constraint-based RL to map-less navigation.
We provide an extensive evaluation of the training and navigation performance in simulation and on a robotic platform.
Our main contributions are:
\begin{itemize}
    \item a case study for combining \ac{il} and \ac{rl}\footnote{Our source code is available here: \url{https://github.com/ethz-asl/rl-navigation}} for map-less navigation
    \item a model for map-less end-to-end motion planning that generalizes to unseen environments
    \item an extensive evaluation of training and generalization performance to unseen environments
\end{itemize}
% This paper is structured as follows:
% In Section~\ref{sec:related-work}, we present the related work.
% Section~\ref{sec:approach} presents the problem and our proposed approach.
% In Section~\ref{sec:experiments} we show our experiments and evaluations before the conclusions are drawn in Section~\ref{sec:conclusion}.

%%%%%%%%%%%%%%%%%%%%%%%%%%%%%%%% Related work %%%%%%%%%%%%%%%%%%%%%%%%%%%%%%%%%%%%%%%%%%%%%%%

\section{Related work}
\label{sec:related-work}
% In this section we give an overview of the existing work in the areas of learning by demonstration and \ac{rl} in connection to the navigation problem with a focus on end-to-end techniques.
% Furthermore, we present the most related approaches combining \ac{il} and \ac{rl} for other applications.

\subsection{Learning by demonstration}
\label{subsec:related-work-il}
Learning by demonstration can be split in two main areas: (i) \ac{irl}, where a reward function is inferred from expert demonstrations and a policy is derived by optimizing this reward with optimal control techniques and (ii) \ac{il}, where expert demonstrations are used to directly infer a policy. Abbeel\etal~\cite{Abbeel2008} present an \ac{irl}-based approach where they teach an autonomous car to navigate in parking lots by observing human demonstrations.
Similarly, Pfeiffer\etal~\cite{pfeiffer2016iros} and Kretzschmar\etal~\cite{kretzschmar2016social} present approaches for navigation in dynamic environments based on \ac{irl}.
By observing pedestrian motion, a probability distribution over pedestrian trajectories is found.
For path planning, the trajectory with the highest probability according to the learned model is chosen with the goal of a close imitation of pedestrian motion.
Wulfmeier\etal~\cite{wulfmeier2016watch} present a similar approach using deep \ac{irl} instead of a combination of classical features in order to learn how to drive an autonomous car through static environments.

%All the above mentioned approaches rely on \ac{irl} techniques which require further optimization during deployment.
In the following, we give an overview of the literature on map-less navigation using \ac{il}.
Muller\etal~\cite{muller2005off} present an image-based approach for end-to-end collision avoidance using imitation learning.
In their work, the focus is on feature extraction and on generalization to new situations.
The overall navigation performance of such approaches is not analyzed.
Another approach focused on perception
is presented by Chen\etal~\cite{chen2015deepdriving}.
They combine learning-based feature extraction using \acp{cnn} with a classical driving controller for an autonomous car.
However, they focus on a lane-following application and do not deal with target-driven navigation.
Kim\etal~\cite{kim2015deep} present an \ac{il} approach for hallway navigation and collision avoidance for an unmanned aerial vehicle (UAV).
They show a working model on a real-world platform, yet the environmental setup is relatively easy and no real navigation capabilities are required.
Sergeant\etal~\cite{sergeant2015autoencoders} present an end-to-end approach for laser-based collision avoidance for ground vehicles demonstrated in simulation and real-world tests.
However, the approach is limited to collision avoidance and cannot be used for target-driven navigation.
Ross\etal~\cite{ross2011dagger} present the Dataset Aggregation (DAGGER) method which collects demonstrations according to the currently best policy but can also query additional expert demonstrations  in order to alleviate the distribution mismatch problem.
%This approach can also be extended to online learning.
One application of the DAGGER algorithm is presented in \cite{ross2013learning}, where  directional commands for forest navigation and collision avoidance are learned from expert demonstrations.
In addition, Kuefler\etal~\cite{kuefler2017imitating} presented an approach based on \ac{gail} \cite{ho2016generative}, where they learn driver models for an autonomous car based on expert demonstrations.
Tai \etal~\cite{tai2017socially} recently applied \ac{gail} to model interaction-aware navigation behavior.
Although conceptually \ac{gail} generalizes better than standard behavioral cloning techniques, it is still constrained by the provided expert demonstrations.

The method we introduce builds upon prior work presented in \cite{pfeiffer2017perception}, where a global planner is used to generate expert demonstrations in simulation.
Given demonstrations, an end-to-end navigation policy mapping from 2D laser measurements and a relative goal position to motion commands is found.
The main drawbacks of this approach are the generalization to new environments --- also due to the specific \ac{cnn} model structure --- and the behavior in situations which were not covered in the training data.% (distribution mismatch).

\subsection{Reinforcement learning}
\label{subsec:related-work-rl}
Bischoff\etal~\cite{Bischoff2013HierarchicalRL} use ideas from hierarchical \ac{rl} to decompose the navigation task in motion planning and movement execution and thus are able to improve the sample efficiency of plain \ac{rl}.
Yet global map information is always assumed to be known.
Zuo\etal~\cite{Zuo2014ARL} use a popular model-free \ac{rl} algorithm, Q-learning, to teach a robot a policy to navigate through a simple spiral maze from sonar inputs only.
% However, no real target information has to be taken into account.
Mirowski\etal~\cite{mirowski2016learning} use auxiliary tasks such as depth prediction and loop closure assessment to improve the learning rate of A3C \cite{mnih2016asynchronous} for simulated maze navigation from RGB images.
Bruce\etal~\cite{Bruce2017OneShotRL} use interactive experience replay to learn how to navigate in a known environment to a fixed goal from images by traversing it only once.
The method presented in \cite{Zhang2017DeepRL} focuses on efficient knowledge transfer across maps and conditions for an autonomous navigation task.
To this end, it uses a particular parametrization of the Q-function, known as successor representation, that decouples task specific knowledge from transferable knowledge.
Zhu\etal~\cite{zhu2017target} present an end-to-end vision-based navigation algorithm that uses the target as an additional input to the policy to learn to achieve proper target-driven navigation.

Chen \etal~\cite{chen2017socialdrl} presented a \ac{rl} approach for collision avoidance in dynamic environments.
Similar to our work, prior demonstrations are used for pre-training, yet their focus lies on learning interactions between multiple agents and the algorithm is not designed for navigation scenarios.
The method presented by Tai\etal~\cite{tai2017virtual} is the most closely related to ours.
In their work, the Asynchronous Deep Deterministic Policy Gradients (ADDPG) algorithm is used to learn a policy from range findings to continuous steering commands for both simulated and real-world map-less navigation tasks.
However, using ADDPG, no collision constraints can be enforced and the models are trained from scratch.
When moving towards real world applications and eventually \ac{rl} training on real platforms, safety and training speed become decisive factors.
Therefore, compared to \cite{tai2017virtual}, we use prior demonstrations for pre-training and \ac{cpo} during \ac{rl} training, targeting the real-world applicability of \ac{rl} approaches.

As experiments in robotics usually require large amounts of time, the problem of reducing the sample complexity of \ac{rl} based approaches has received increasing attention recently.
Using a combination of \ac{il} and \ac{rl} to obtain a sample efficient and robust learning algorithm has previously been explored in robotics in the context of manipulation tasks \cite{Balaguer2011CombiningIA,Zhu2018ReinforcementAI}.
In this context, the main challenge consists in using human demonstrations that may not be replicable by the robot due its dynamics.
In the case of navigation, this is usually not a concern.
However, navigation tasks present challenges in terms of safety.
Even small deviations from the expert policy may lead to a crash.
To the best of our knowledge, our method is the first to use expert demonstrations to boost \ac{rl} learning performance in the context of map-less autonomous navigation.

%%%%%%%%%%%%%%%%%%% Approach %%%%%%%%%%%%%%%%%%%%%%%%%%

\section{Approach}
\label{sec:approach}

\subsection{Problem formulation}
\label{subsec:problem-formulation}
Classical path planning techniques \cite{lavalle2006planning} require prior knowledge of the environment for navigation.
In case of unknown or constantly changing and dynamic environments, obtaining and maintaining an accurate map representation becomes increasingly difficult or even unfeasible.
Therefore, map-less navigation skills based solely on local information available to the robot through its sensors are required.

Given the sensor measurements $\y$ and a relative target position $\g$, we want to find a policy $\pi_{\params}$ parametrized by $\params$ which maps the inputs to suitable control commands, $\u$,
i.e.
\begin{equation}
    \u = \pi_{\params}(\y, \g).
\end{equation}
The required control commands are comprised of the translational and rotational velocity.
As the mapping from local sensor and target data to control commands can be arbitrarily complex, learning how to plan from experience in an end-to-end fashion using powerful non-linear function approximators, such as neural networks, has become more prominent. % within the last decade.
% To this end, typically two different types of machine learning techniques are applied: \acf{il} and \acf{rl}.
%While the former is sample efficient, it suffers from overfitting and poor generalization to unseen scenarios.
%The latter, despite its robustness to previously unseen scenarios, notoriously requires a large amount of data to achieve satisfactory navigation performance.
In this work, we aim at combining \ac{il} and \ac{rl} to obtain a sample efficient and robust learning based navigation algorithm.
We do this in a sequential fashion by using the result from \ac{il} to initialize our \ac{rl} method.
In the remainder of this section we introduce separately the underlying neural network model, the \ac{il} and \ac{rl} components of our method.

\subsection{Neural network model}
\label{subsec:nn-model}
\begin{figure}[t]
\centering
    \includegraphics[width=0.9\columnwidth,trim=0 15 0 0]{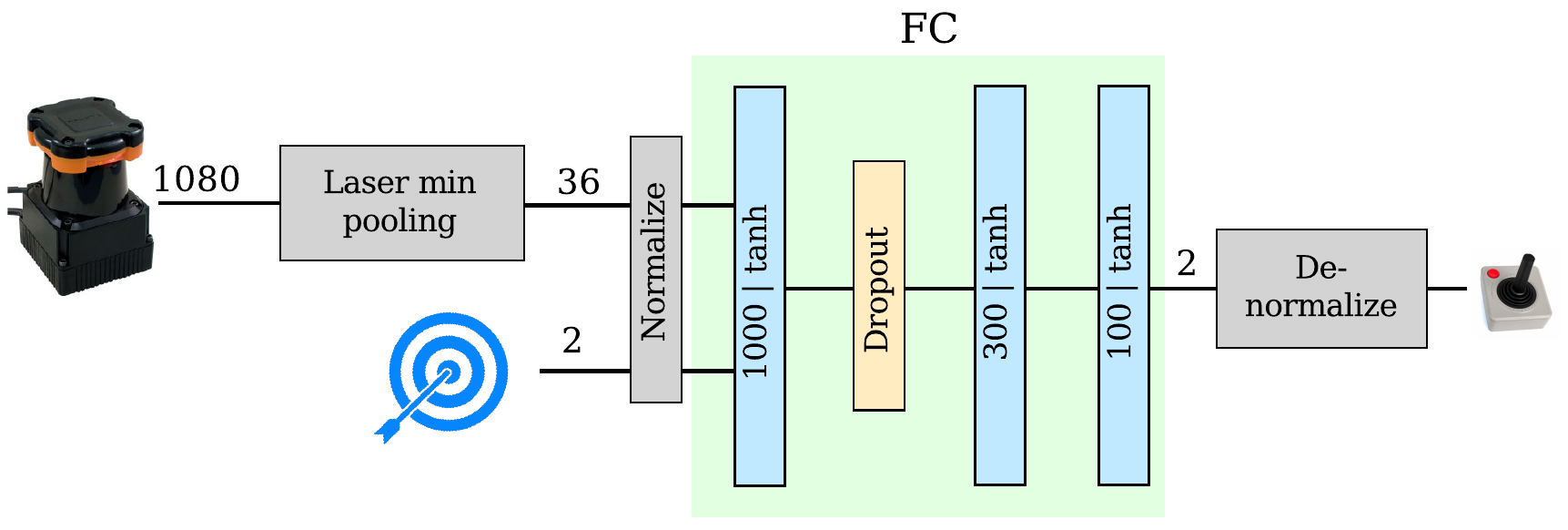}
    \caption{The neural network model for $\pi_{\params}$.
    The normalized input data is fed through three fully connected layers with tanh activation functions.
    Between layer one and two, dropout is added during \ac{il} training.
    The outputs are de-normalized to obtain physical control commands from the neural network.}
    \label{fig:nn-structure}
\vspace{-4mm}
\end{figure}

The neural network model which represents $\pi_{\params}$, is shown in Figure \ref{fig:nn-structure}.
In this work, the inputs to the model are 2D laser range findings and a relative target position in polar coordinates w.r.t. the local robot coordinate frame.
In contrast to \cite{pfeiffer2017perception}, where a \ac{cnn} was used to extract environmental features, this model is simplified and only relies on three fully connected layers.
While the \ac{cnn} allows to find  relevant environmental features, we found that it tends to overfit to the shapes of the obstacles presented during training.
Instead, we use minimum pooling of the laser data and compress the full range of 1080 measurements into 36 values, where each pooled value $\y_{p,i}$ is computed as:
\begin{equation}
    \y_{p,i} = \min \big ( \y_{i \cdot k}, \dots, \y_{(i+1) \cdot k - 1} \big ),
\end{equation}
where $i$ is the value index and $k$ is the kernel size for 1D pooling.
In our case, we chose $k=30$.
Using min-pooling, safety can be assured, yet detailed environmental features may get lost.
The resulting simplified neural network model can be trained more efficiently and is less likely to overfit to specific obstacle shapes. % , which was one problem found in \cite{pfeiffer2017perception}.
Furthermore, the inputs are normalized before being fed to the neural network model.
The pooled laser measurements are cropped and then mapped to lie in the interval $[-1, 1]$ by applying the normalization
$2 \cdot \big(1 - \frac{\min(\y_{p,i}, r_{\text{max}})}{r_{\text{max}}}\big) -1$, where $r_{\text{max}}$ is the maximum laser range.
The same normalization is applied to the relative target position.
The outputs of the neural network, which also lie in the interval $[-1,1]$, are de-normalized and mapped to translational and rotational velocities.
%For \ac{il} training, we introduce a 50\% dropout.

%The presented model does not use any external state feedback or internal memory.
%It is purely based on feed-forward computations.

\subsection{Supervised pre-training via behavior cloning}
\label{subsec:pretraining}
%In order to improve the performance and sample complexity of the succeeding \ac{rl}, the policy is pre-trained using supervised \ac{il} based on expert demonstrations (also referred to as behavioral cloning), similar to \cite{pfeiffer2017perception}.
In order to improve the performance and sample complexity of the succeeding \ac{rl}, the policy is pre-trained using supervised \ac{il} based on expert demonstrations similar to \cite{pfeiffer2017perception}.
The goal is to imitate the expert as closely as possible, given the representation limitations of the neural network model.
Compared to plain \ac{il}, where the performance of the final model is limited by the performance of the expert demonstrations, \ac{ril} can overcome this limitation through self-improvement.
% The output of the \ac{il} policy is deterministic and no further sampling actions are conducted.

\subsection{Reinforcement learning}
\label{subsec:reinforcement-learning}
\subsubsection{Background information}

%A policy, $\pi_{\theta}$ is a mapping from observations to actions. The goal of Reinforcement Learning(RL) is to find a policy that maximizes the expected sum of discounted rewards.
Given a \ac{mdp}, $M=\langle \mathcal{S}, \mathcal{A}, \mathcal{P}, \mathcal{R}, \gamma\rangle$, where $\mathcal{S}$ is the state space, $\mathcal{A}$ is the action space, $\mathcal{P}(\cdot|s_t, a_t):\mathcal{S} \times \mathcal{S} \times \mathcal{A}:\rightarrow \mathbb{R}_+$ is the transition probability distribution, $\mathcal{R}(\cdot, \cdot):\mathcal{S}\times \mathcal{A}\rightarrow \mathbb{R}$ is the reward function and $\gamma \in [0, 1]$ is the discount factor, \ac{rl} aims to find a policy $\pi_{\theta}$, mapping states to actions and parametrized by $\params$, that maximizes the expected sum of discounted rewards,
\begin{equation}\label{eq:discountedR}
    J(\params) = \mathbb{E} \bigg[\sum_{t=0}^{T} \gamma^t \cdot r(s_t, \pi_{\theta}(s_t)) \bigg],
\end{equation}
where $T$ is the time horizon of a navigation episode.
In our case, $s_t$ consists of laser measurements and the target information, $a_t$ of the control commands.

Policy gradient methods \cite{williams1992simple} are model-free \ac{rl} algorithms that use modifications of stochastic gradient descent to optimize $J(\params)$ with respect to the policy parameters $\params$.
%Policy gradient methods are model-free methods and thus they can be used in scenarios where the dynamic models of the environment are unavailable.
However, they suffer from a high variance in gradients, resulting in undesirably large updates to the policy.
A popular technique to reduce model variance and ensure stability between updates is \ac{trpo} \cite{schulman2015trust}.
To this end, it restricts the change in policy at each update by imposing a constraint on the average \ac{kl} divergence between the new and old policy.

% \begin{itemize}
%     \item Introduction of a baseline function in the computation of the gradient estimate so that variance is reduced without introducing bias. Different choices for this baseline function are discussed in \cite{peters2008reinforcement}.
%     \item Generalized advantage estimation \cite{schulman2015high} introduces an additional parameter to use a weighted average of k-step advantage estimators.
%     This significantly reduces the variance of the gradient estimator at the cost of some bias.
%     \item \ac{trpo} \cite{schulman2015trust} restricts the change in policy at each update by imposing a constraint on the average \ac{kl} divergence between the new policy and old policy.
%     This results in reliable and efficient exploration and stable policy updates.
% \end{itemize}

Enforcing safety is crucial when dealing with mobile robotics applications.
Often, safety in RL is encouraged by imposing high cost on unsafe states.
However, this requires tuning such cost.
If it is too low, the agent may decide to experience unsafe states for short amounts of time as this will not severely impact the overall performance (Eq. \ref{eq:discountedR}).
Conversely, if the cost is too high, the agent may avoid exploring entire portions of the state space to avoid the risk of experiencing unsafe states.
A more elegant and increasingly popular way of ensuring safety in \ac{rl} is to treat it as a constraint \cite{achiam2017cpo,Berkenkamp2017SafeMR}.
In particular, in this work, we use a safety constrained extension of \ac{trpo} known as \acf{cpo} \cite{achiam2017cpo} to ensure safety.
Given a cost function $C:\mathcal{S}\times \mathcal{A}:\rightarrow \mathbb{R}$, let $J_C(\theta)$ indicate the expected discounted return of $\pi_\theta$ with respect to this cost
\begin{equation}\label{eq:discountedC}
    J_C(\params) = \mathbb{E} \bigg[\sum_{t=0}^{T} \gamma^t \cdot C(s_t, \pi_{\theta}(s_t)) \bigg].
\end{equation}
\ac{cpo} finds an approximate solution to the following problem,
\begin{equation}\label{eq:cMDP}
    \params^*=\textrm{arg max}J(\params),~ s.t.~ J_C(\params)\leq \alpha.
\end{equation}
%As a consequence, it is the algorithm of choice for the \ac{rl} part of our approach.

\subsubsection{Training process}
For training, the neural network model is first initialized either randomly (pure \ac{rl}) or using \ac{il} (\ac{ril}).
We use a stochastic policy where the actions are sampled from a 2D Gaussian distribution having the de-normalized values of the output of the neural network as mean, and a 2D standard deviation which is a separate learn-able parameter.
Using a supervised \ac{il} model thus only influences the initialization of the \ac{rl} policy.
During training we randomly select a start and target position and collect robot experience samples by running an episode using the current policy $\pi_{\params}$ for a fixed number of time steps or until the robot reaches the target.
At each policy update, we use a batch of samples collected from multiple episodes.

%\color{red}
The agent's objective is to learn to reach the target in the shortest possible number of time-steps while avoiding collisions with surrounding obstacles.
The reward function provides the required feedback to the robot during the learning process.
In this work, we investigate different choices for the reward function encoding various degree of information about the task.
These rewards can be expressed by:
%The desired behavior to successfully reach the target is encouraged by giving a positive reward when the robot reaches the target.
%However, having a sparse reward for final success makes the learning process difficult as the agent struggles to differentiate which actions had a positive or negative effect toward the accomplishment of the task.
%\todomark{add credit assignment problem}
%Therefore, the reward function is shaped to provide continuous feedback for each action by rewarding/penalizing the agent for getting closer/further to/from the goal from the current location along the shortest feasible path.
%Let $d(s)$ denote the distance from $s$ to the goal along the shortest feasible path to the target, which takes into account the position of obstacles in the map and is computed using the Dijkstra algorithm \cite{dijkstra1959note}.
%The combined reward function is given by:
\[r(s_{t}) =
\begin{cases}
    10,& \text{if success}\\
    -(d(s_{t}) - d(s_{t-1})),& \text{otherwise.}
\end{cases}\]
Setting $d(s)=0,~\forall s \in \mathcal{S}$ we encode the minimum information required to carry out the task.
This \textit{sparse reward} makes the learning process difficult due to the credit assignment problem, i.e. the fact that all the actions taken in an episode get credit for its outcome regardless of whether they contributed to it or not.
An alternative to such choice is to set $d(s)$ to the \textit{Euclidean distance} between $s$ and the target.
This reward provides continuous feedback for each action by rewarding/penalizing the agent for getting closer/further to/from the goal in Euclidean space.
However, it does not consider the placement of obstacles in the environment.
The last option we investigate consists in setting $d(s)$ to the distance between $s$ and the goal along the \textit{shortest feasible path} that can be computed using the Dijkstra algorithm.
Note, the agent does not have any knowledge about $d(\cdot)$.
%\color{black}
This distance is only used to compute the reward which the agent receives from the environment during training.

Using a negative reward for collisions makes the policy highly sensitive to this reward's magnitude, resulting in a delicate trade-off between two different objectives --- reaching the target and avoiding crashes.
However, in constrained \ac{mdp}s, we can encode collision avoidance through a constraint on the expected number of crashes allowed per episode. Let $\mathcal{S}_c \subset \mathcal{S}$ denote the set of states that correspond to a crash. We define a state dependend cost function as follows:
\begin{equation}\label{eq:cost}
    c(s_t)=\mathcal{I}(s_t \in \mathcal{S}_c),
\end{equation}
where $\mathcal{I}$ is the indicator function.
In our experiments, we noticed the robot stays in a crash state for four consecutive timesteps on average.
By setting the discount factor for the cost --- which does not have to be equal to the one for the reward --- close to 1 and introducing the constraint value $\alpha$, we can constrain the total number of expected crashes per episode to be approximately less or equal to $\frac{\alpha}{4}$.
In our model we set $\alpha=0.4$.
This value was found empirically by testing values between 0.0 and 0.6 in a simple environment.
While training, we allow for multiple crashes in each episode.
This leads to more crash samples in the training set and makes it easier to reach the target, thus making the training process more efficient.

%%%%%%%%%%%%%%%%%%%%%%%%%%%% Experiments %%%%%%%%%%%%%%%%%%%%%%%%%%%%%%%%%%%

\section{Experiments}
\label{sec:experiments}
This section presents the experiments conducted in simulation and on the real robotic platform.
The goal of the experiments is to investigate the influence of pre-training the \ac{rl} policy, to compare constraint-based to fixed penalty methods and analyze the influence of the reward functions presented in Section~\ref{sec:approach}.
We also compare to models presented in prior work \cite{tai2017virtual}.
Furthermore, we investigate the generalization performance of the navigation policies to unseen scenarios and the real world, which is also shown in our video\footnote{\url{https://youtu.be/uc386uZCgEU}}.
Our work does not intend to show that we can outperform a global graph-based planner in known environments, where graph-based solutions are fast and can achieve optimal behavior.
The goal of our experiments is to investigate the limits of motion planning with local information only.

\subsection{Experimental setup}
\label{subsec:experimental-setup}
The models are purely trained in simulation since it is a safe, fast and efficient way of training and evaluating the model.
Additionally, there are no physical space constraints and the environment structure can be changed almost arbitrarily.
% Moreover, the trained model may not have the desired behavior, due to mis-specification of the reward function or data inefficiency \cite{amodei2016concrete}.
% The agent learns collision avoidance through its own experiences.
Models trained in simulation have previously been shown to successfully transfer to the real-world  \cite{tai2017virtual,zhu2017target,pfeiffer2017perception}.

The experiments are based on a differential drive Kobuki TurtleBot2\footnote{\url{http://kobuki.yujinrobot.com/about2}} platform equipped with a front-facing Hokuyo UTM laser range finder with a field of view of \SI{270}{\degree}, maximum range of \SI{30}{\meter} and 1080 range measurements per revolution.
For on-board computations we resort to an Intel\textsuperscript{\tiny\textregistered} NUC with an i7-5557U processor and without any GPU, running Ubuntu 14.04 and ROS \cite{ros} as a middleware.
The motion commands are published with a frequency of \SI{5}{\hertz}.

\begin{figure}
\centering
\begin{subfigure}{0.18\columnwidth}
\centering
    \stackunder[3pt]{\includegraphics[width=\linewidth,trim=0 0 0 0]{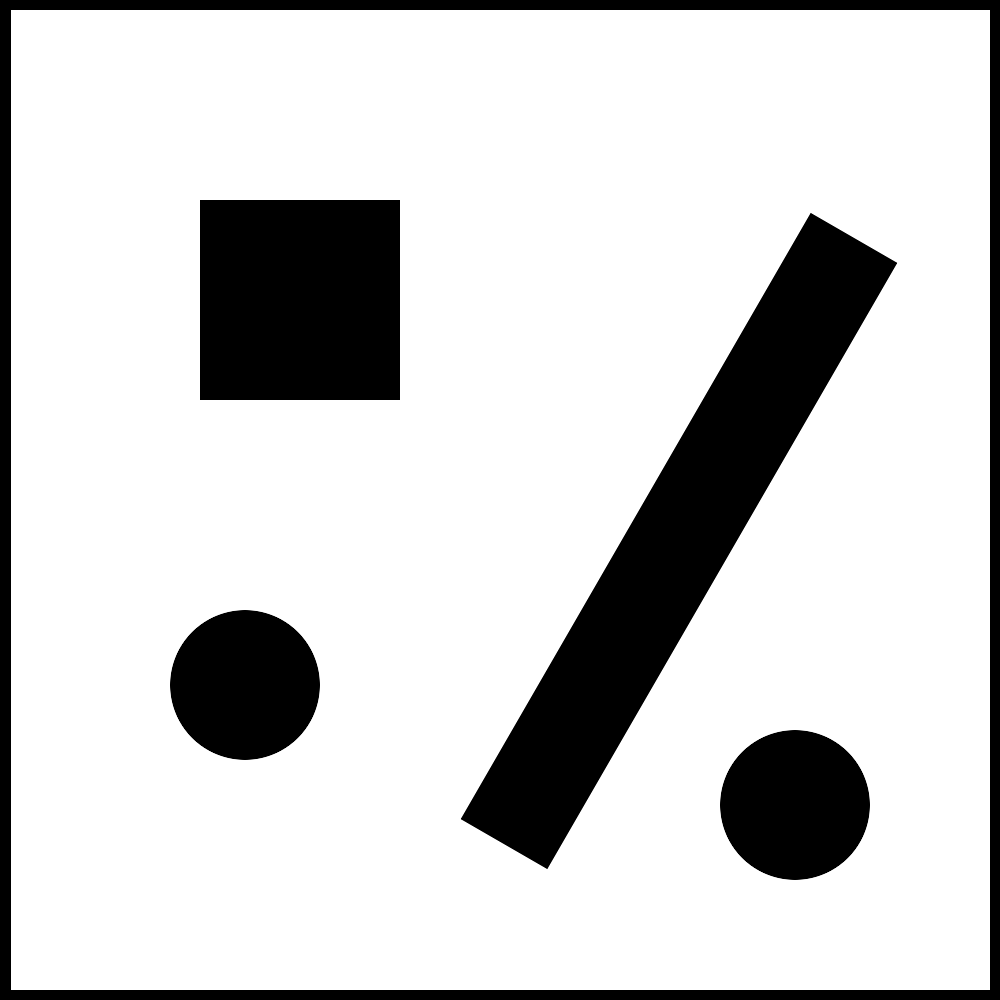}}{\textit{\tiny{simple}}}
    \label{fig:map-simple}
\end{subfigure}
\begin{subfigure}{0.18\columnwidth}
\centering
    \stackunder[3pt]{\includegraphics[width=\linewidth,trim=0 0 0 0]{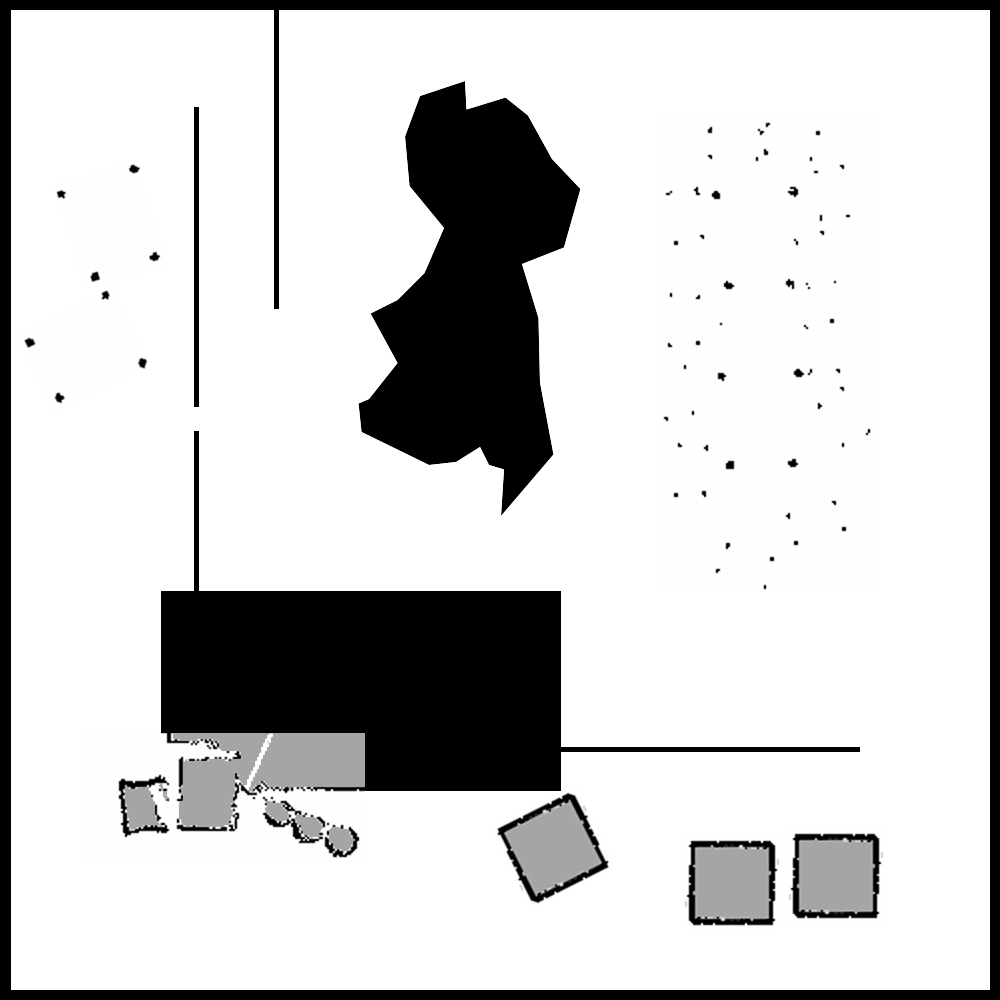}}{\textit{\tiny{complex}}}
    \label{fig:map-complex}
\end{subfigure}
\begin{subfigure}{0.18\columnwidth}
\centering
    \stackunder[3pt]{\includegraphics[width=\linewidth,trim=0 0 0 0]{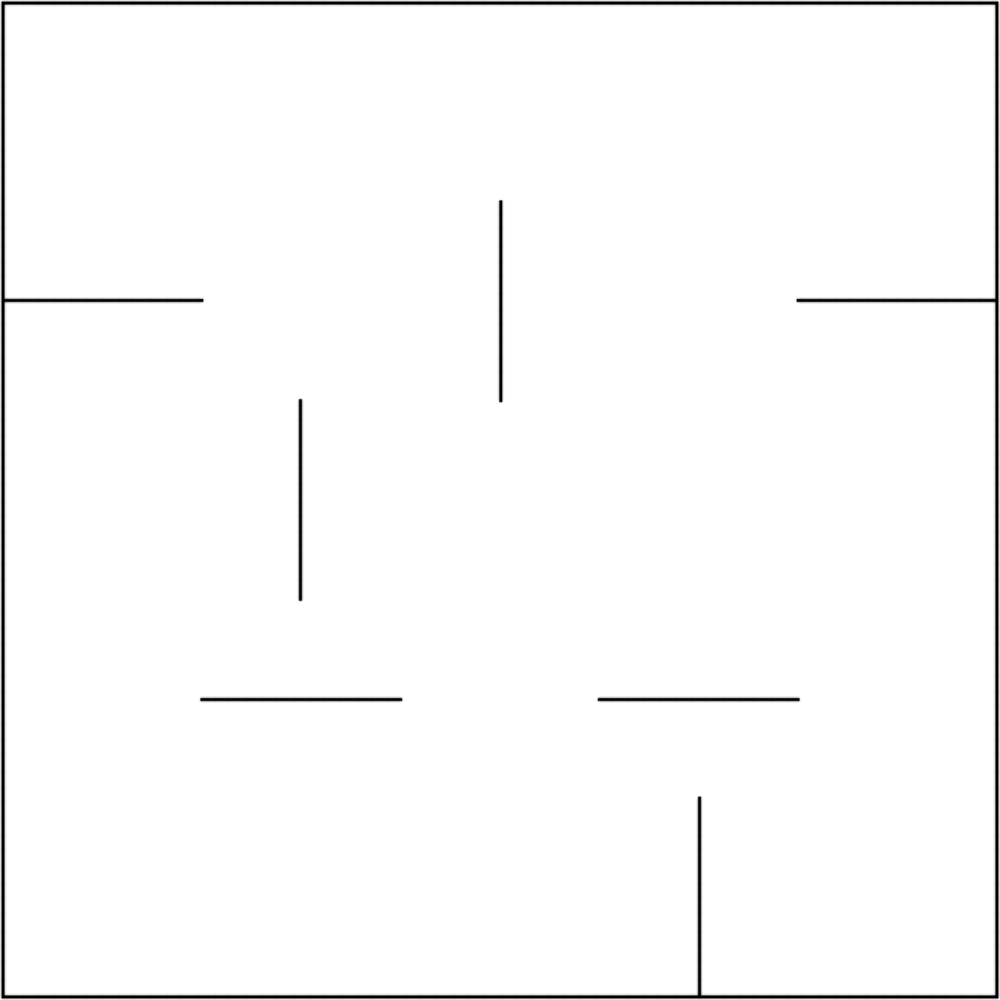}}{\textit{\tiny{TM-1}}}
    \label{fig:map-rl-1}
\end{subfigure}
\begin{subfigure}{0.18\columnwidth}
\centering
    \stackunder[3pt]{\includegraphics[width=\linewidth,trim=0 0 0 0]{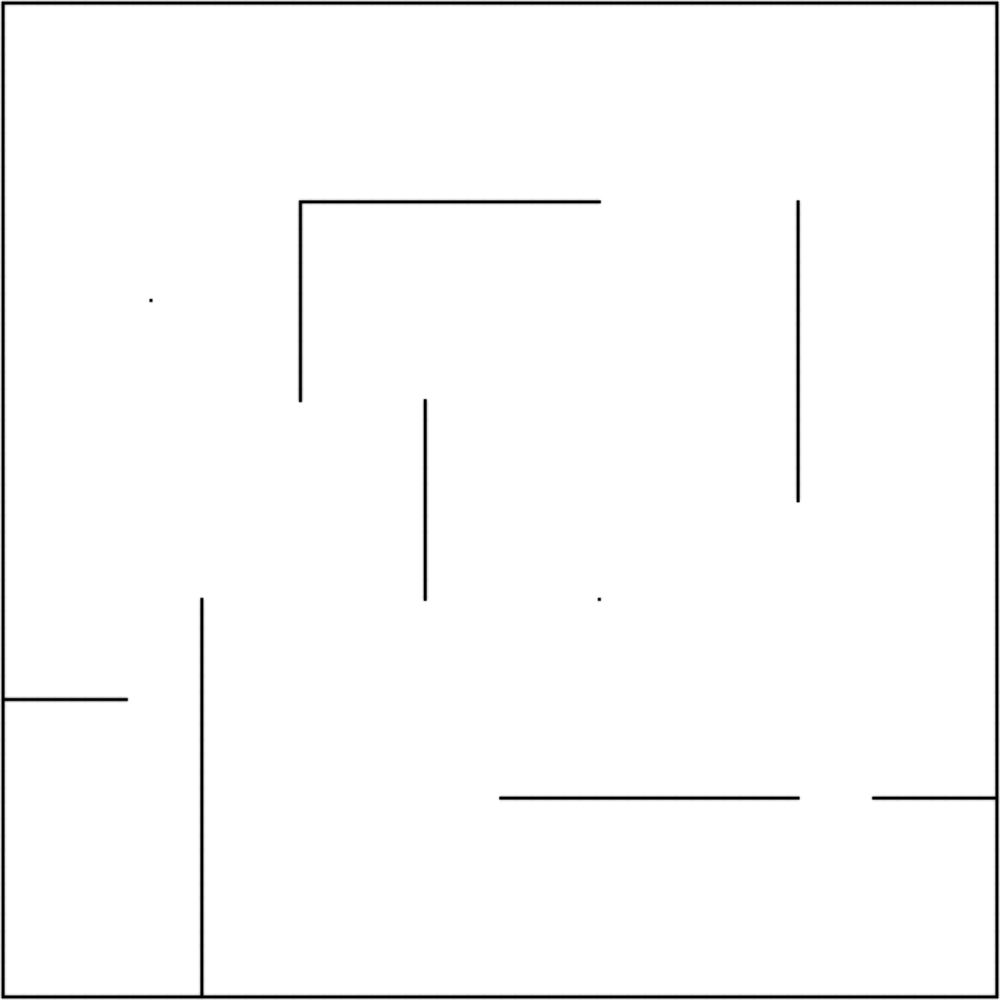}}{\textit{\tiny{TM-2}}}
    \label{fig:map-rl-2}
\end{subfigure}
\begin{subfigure}{0.18\columnwidth}
\centering
    \stackunder[3pt]{\includegraphics[width=\linewidth,trim=0 0 0 0]{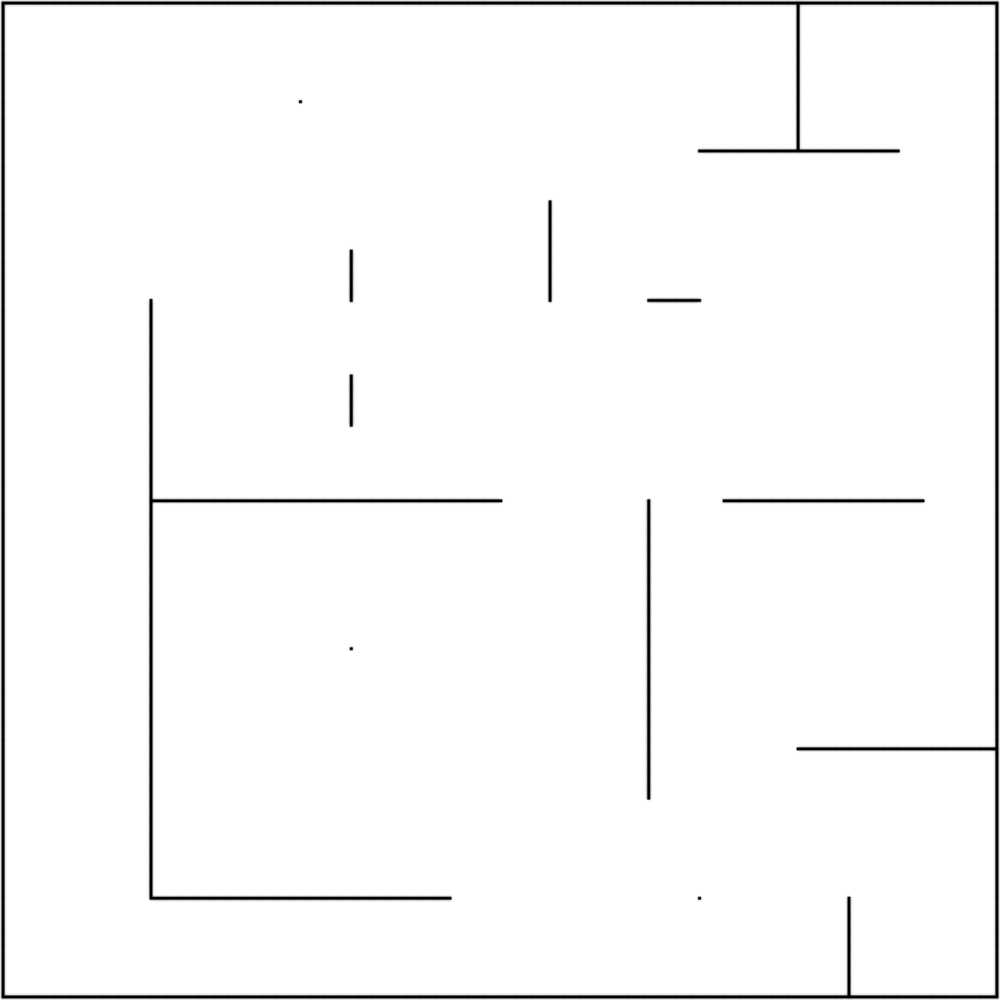}}{\textit{\tiny{TM-3}}}
    \label{fig:map-rl-3}
\end{subfigure}
\vspace{-3mm}
\caption{Training maps for \ac{il} and \ac{rl}.
The \textit{TM} vary significantly in difficulty.
Maps can be better viewed by zooming in on a computer screen.}
\label{fig:training-maps}
\vspace{-6mm}
\end{figure}

\subsection{Model training}
Different procedures for model training are applied: (i) pure \ac{il}, (ii) pure \ac{rl} and (iii) \ac{ril}, which is a combination of both.
In order to test the influence of the complexity and the diversity of the training environments on test performance, we train the models on five maps (or subsets of them) as shown in Figure~\ref{fig:training-maps}.
The pure \ac{il} models are trained in the \textit{simple} and \textit{complex} maps, the \ac{rl} part is conducted on all three \textit{TM} maps.
% One exception is RL1, which is purely trained on \textit{TM1}.
% Most of the \ac{il} models are trained either on the \textit{simple} or \textit{complex} map, while the \ac{rl} training happens on the maps \textit{TM-1} to \textit{TM-3}.
% One \ac{il} model is also trained on the \textit{TM} maps in order to investigate the effect of an environment change during training.
%As a quantitative measure for the difficulty of the maps we use the ratio between euclidean distance and actual path distance between randomly sampled points.
Similarly, for \ac{ril}, \ac{il} is conducted on the \textit{simple} and \textit{complex} maps and the \ac{rl} part takes place on the \textit{TM} maps.
We do this separation in order to investigate how demonstrations from a different environment can be transferred to the \ac{rl} training.

The expert demonstrations used for \ac{il} are generated using the ROS \texttt{move\_base}\footnote{\url{http://wiki.ros.org/move_base}} navigation stack to navigate between random start and target positions, as presented in \cite{pfeiffer2017perception}.
We use an expert planner instead of a human to make the demonstrations more consistent and time efficient.
We note that the demonstrations are suboptimal for \ac{rl}, as they are generated based on a different cost function and also in a different environment.
After recording the demonstrations, one IL training iteration takes around \SI{7}{\milli \second} on an Intel\textsuperscript{\tiny\textregistered} i7-7700K processor and a Nvidia GeForce GTX 1070 GPU.
Therefore, IL model training takes between one hour (s$_{10}$, 500\,k iterations) and around 2.5 hours (c$_{1000}$, 1.5\,M iterations).

Table \ref{tab:models} summarizes all the models we trained.
% For each model, it indicates the number of expert trajectories, the demonstration environment and the RL training procedure (\ac{cpo} vs. \ac{trpo}) and the reward signal, if applicable.
% As Table \ref{tab:models} suggests, the expert demonstrations range from only 10 trajectories on the \textit{simple} map up to 1000 trajectories on the \textit{complex} map.
Our case study presents constraint based \ac{ril} yet compares to a broad range of different models:
We vary the number of demonstrations (from 10 to 1000), the RL training procedure (CPO, TRPO) and reward signals (sparse, Euclidean and shortest distance) in order to provide insights into how those factors influence map-less navigation.
The TRPO training procedure is the fixed collision penalty version of CPO with a collision constraint, as described in Section \ref{sec:approach}.

\vspace{2mm}
\begin{table}[tbp]
\caption{Model details, including the maps and number of trajectories used for \ac{il} and the reward signal used for \ac{rl}.
Besides CPO1, all models are trained on all three $TM$ maps.
CPO and TRPO in the model name specify the \ac{rl} training procedure, the subscript of TRPO indicates the fixed penalty weight for collisions.}
\label{tab:models}
\vspace{-3mm}
\begin{center}
\setlength\tabcolsep{4 pt}
\begin{tabular}{|c|c||c|c|c|  }
\hline
  												& model name					 & \ac{il}-map(s) & \#\ac{il} traj. & \ac{rl} reward  \\
 \hline
  \multirow{8}*{\ac{ril}} & s$_{10}$+CPO$_{sparse}$ 	& simple    & 10        & sparse    \\
  												& s$_{10}$+CPO$_{Eucl.}$ 		& simple    & 10        & Euclidean \\
  												& s$_{10}$+CPO		 					& simple    & 10        & short     \\
												  & s$_{1000}$+CPO   					& simple    & 1000      & short     \\
												  & 123$_{1500}$+CPO 					& 1+2+3     & 500 each  & short     \\
												  & c$_{1000}$+CPO   					& complex   & 1000      & short     \\
												  & s$_{1000}$+TRPO$_{c0.1}$	& simple    & 1000      & short     \\
												  & s$_{1000}$+TRPO$_{c1.0}$  & simple    & 1000      & short     \\
\hline
 \multirow{2}*{\ac{il}} & s$_{10}$           & simple    & 10        & ---        \\
 												& c$_{1000}$         & complex   & 1000      & ---        \\
 \hline
 \multirow{4}*{\ac{rl}} & CPO1                & ---       & 0         & short        \\
  											& CPO123              & ---       & 0         & short     	 \\
  											& CPO123$_{sparse}$   & ---       & 0         & sparse  	 \\
  											& TRPO123$_{c0.1}$    & ---       & 0         & short				 \\
 \hline
\end{tabular}
\end{center}
\vspace{-5mm}
\end{table}

%During \ac{rl}, the training environment is uniformly sampled among the three (except for CPO1, where it is only \textit{TM-1}) \textit{TM} maps (see Figure \ref{fig:training-maps}).
During \ac{rl}, the training environment is uniformly sampled among the three \textit{TM} maps (see Figure \ref{fig:training-maps}).
One training iteration --- for which we consider a batch consisting of 60\,k time steps --- takes around \SI{180}{\second} using the accelerated Stage \cite{vaughan2008massively} simulation.
Therefore, 1000 iterations require around 50 hours of training time using the simulation, which is a real-time equivalent of around 100 days.
This further motivates the need to find a good policy initialization by \ac{il} in order to reduce the training time significantly.

\begin{figure}[t]
\centering
    \includegraphics[width=\columnwidth,trim=10 15 0 0]{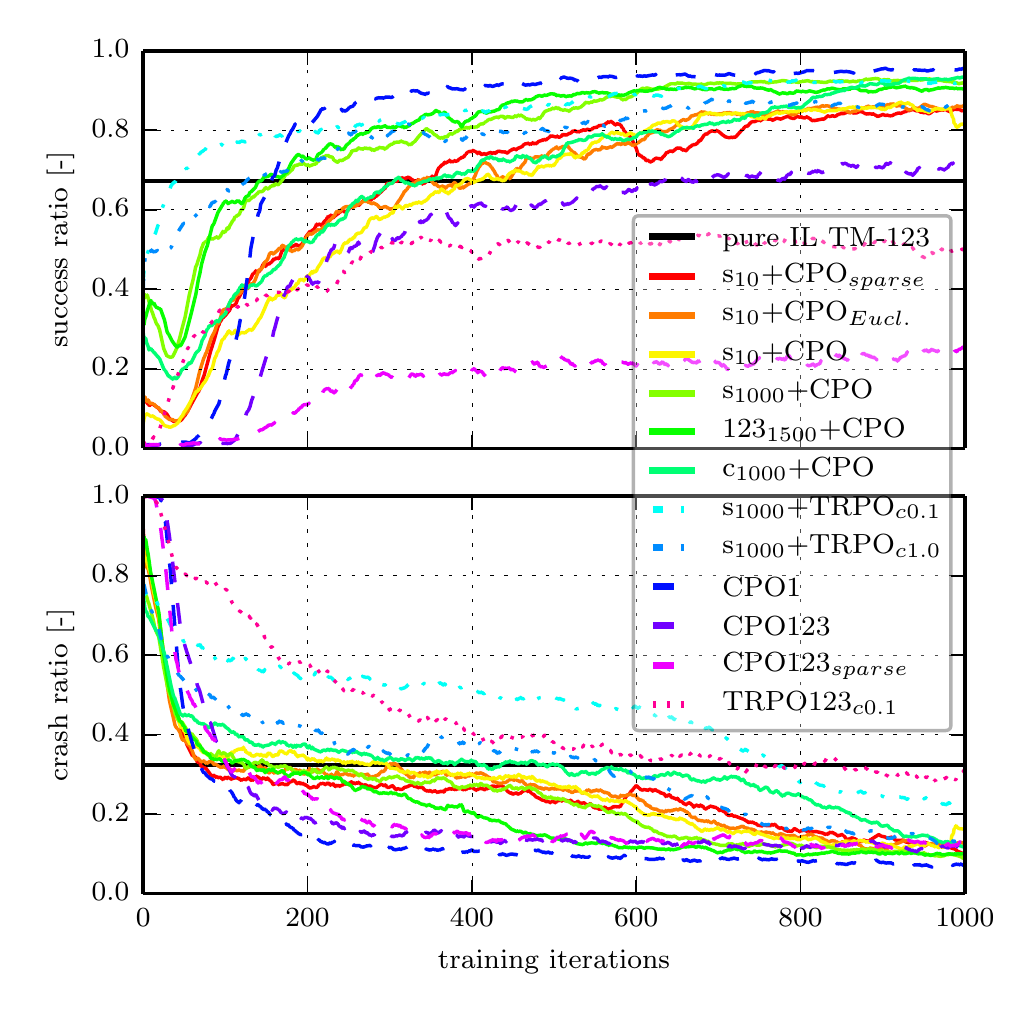}
    \caption{The evolution of navigation success and crash rates throughout the \ac{rl} training process of various models.
    The curves indicate the rolling mean of success and crash rates over 20 steps.
    The models contain pure \ac{rl} models, pure \ac{il} models and \ac{ril} which differ in the amount and complexitiy of pre-training, the reward structure and the RL training procedure.
    The black line indicates the performance of \ac{il} on the training maps (\textit{TM-123}) as a reference.
    For the RL training, where multiple runs were conducted, only the best one is shown.}
    \label{fig:training_curves}
\vspace{-7mm}
\end{figure}

Figure~\ref{fig:training_curves} shows the success and crash rates of a broad range of models during \ac{rl} training alongside the performance of pure \ac{il} trained on all \textit{TM} maps.
% This pure \ac{il} model is not included in Table \ref{tab:models} and is not tested in the testing environment.
This IL model only serves as a baseline to evaluate the progress of the \ac{rl} and \ac{ril} methods during training.
CPO1 differs from all the other models during training as it is exclusively trained on the simplest \textit{TM} map (\textit{TM}1).
However, it will be shown that this model does not generalize well to more complex test environments.
From Figure~\ref{fig:training_curves} the following can be shown:
\subsubsection{Difference between the models which were pre-trained using \ac{il} and the ones based on pure \ac{rl} using \ac{cpo} / \ac{trpo}}
While the pre-trained models already start at a certain success rate (depending on the performance of the \ac{il} model), it takes a significant amount of iterations for the \ac{rl} models to reach the target in the majority of the cases.
Comparing the TRPO and CPO versions of the different models also shows the potential problems of constraint-based methods.
Initially, the cost that defines the safety constraint (Eq.~\ref{eq:cost}), used in \ac{cpo} has very high values and the agent learns to satisfy it.
This also explains the drop in success rate early during training, which all \ac{ril} models trained with CPO have in common.
Therefore, in this phase, the agent learns to avoid crashes and unlearns the behavior of reaching the target, which is also supported by the crash rate curves. % in Figure \ref{fig:training_curves}.
Both models (with high and low cost of collision) trained with TRPO do not show this behavior as no constraint needs to be satisfied.
Therefore TRPO allows for more ``risky'' exploration initially.
This further motivates to use pre-training when using constraint-based RL, as it provides enough intuition to reach the target while the agent can learn how to satisfy the safety constraint.
This would be hard otherwise, as exploration through Gaussian perturbation of a nominal motion command is inherently local in the policy space.
The difference becomes even more pronounced for the simpler reward structures, such as sparse target reward.
While the agent is stuck with a low success rate for CPO123$_{sparse}$ and mostly learns collision avoidance,
pre-training with only 10 demonstrations allows the agent to successfully reach the goal in the vast majority of the cases (s$_{10}$+CPO$_{sparse}$).
With pre-training, sparse and full (shortest path) reward reach about the same final performance.
\subsubsection{Problem of fixed penalty methods}
While, e.g., s$_{1000}$+CPO reaches a high final success rate and a low crash rate, s$_{1000}$+TRPO$_{c0.1}$ reaches similar success rates, yet struggles with significantly more crashes.
On the other side, s$_{1000}$+TRPO$_{c1.0}$ reaches a similar crash rate yet does not achieve the same final success rate.
This difficulty of fixed penalty parameter tuning was already raised in \cite{achiam2017cpo}.
\subsubsection{Final performance is affected by the initial starting state} Models initialized using more complex maps and/or more trajectories not only perform better but also learn faster.
Even a very small amount of demonstrations can significantly improve the overall performance.
The \ac{ril} models reach the final performance of CPO123 after less than one fifth of the iterations ($\approx 200$) as pre-training provides a good initial policy and makes the stochastic exploration more target-aimed.
This confirms our initial hypothesis that the prior \ac{il} can significantly reduce the training time in \ac{rl} applications.

\subsection{Simulation results}
\label{subsec:simulation-results}

\begin{figure}[t]
\vspace{-0mm}
\centering
\begin{subfigure}{0.49\columnwidth}
\centering
    \stackunder[3pt]{\includegraphics[width=0.7\linewidth,trim=0 0 0 0]{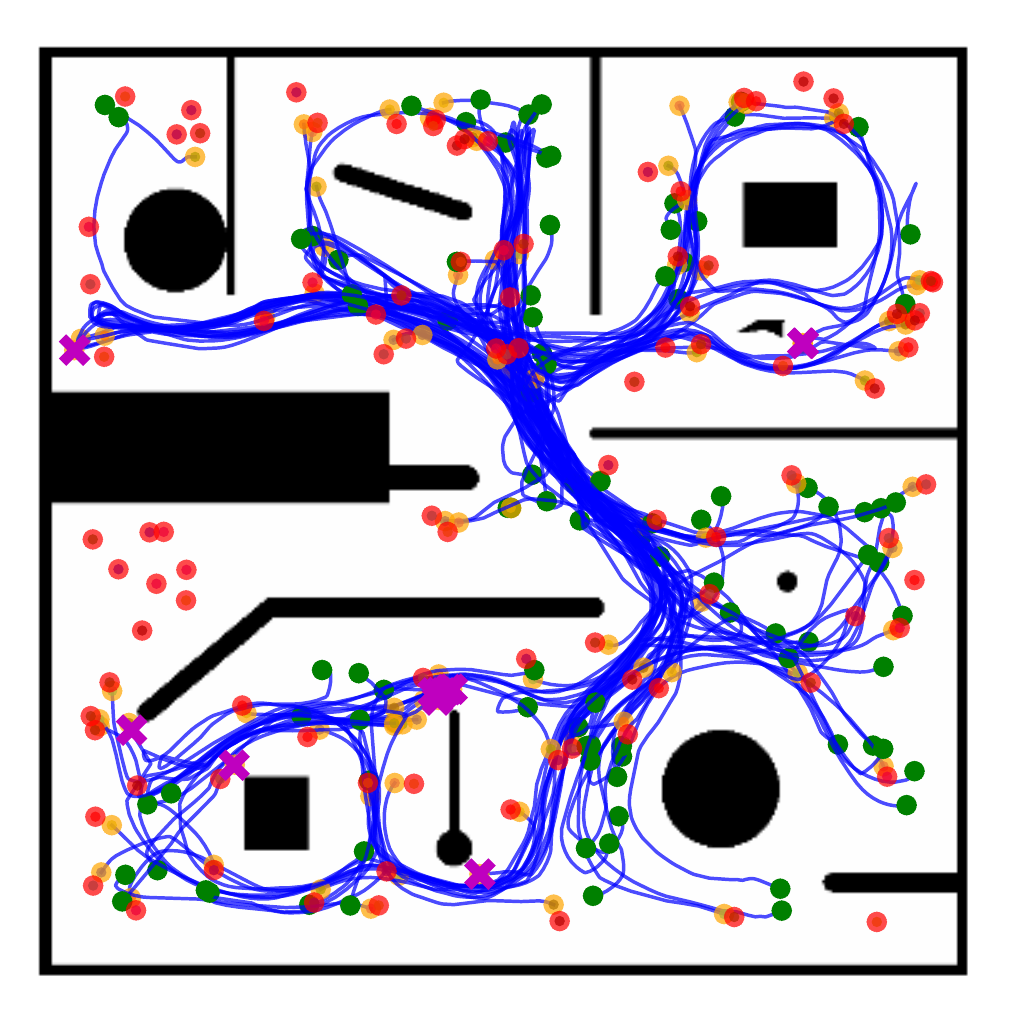}}{\textit{maze}}
    \label{fig:traj-on-map-sim-maze}
\end{subfigure}
\begin{subfigure}{0.49\columnwidth}
\centering
    \stackunder[3pt]{\includegraphics[width=0.7\linewidth,trim=0 0 0 0]{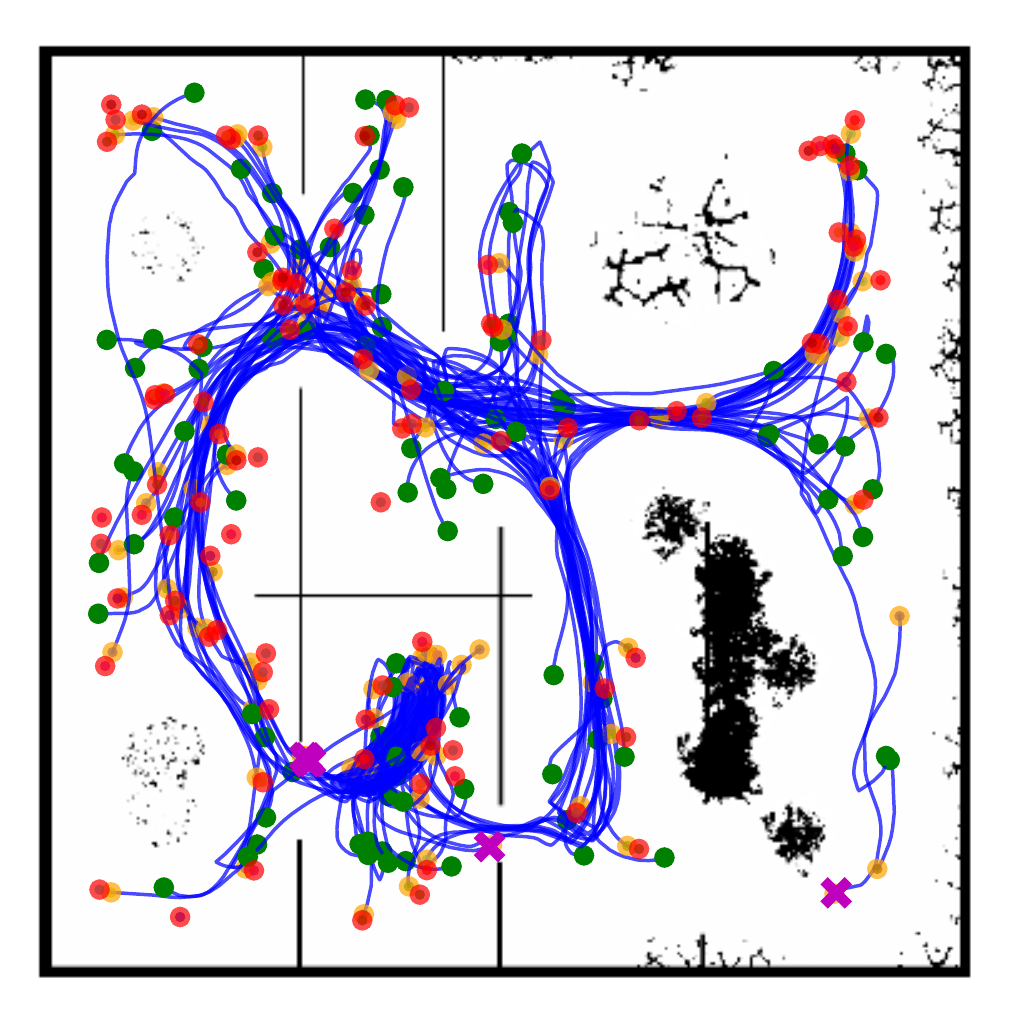}}{\textit{clutter}}
    \label{fig:fig:traj-on-map-sim-blobs}
\end{subfigure}
\vspace{0mm}
\caption{Evaluation runs between 100 randomly sampled start and target positions on the two unknown test maps (both $\SI{10}{\meter} \times \SI{10}{\meter}$).
The model used for visualization is c$_{1000}$+CPO.
The trajectories are shown in \textit{blue}, the starting positions in \textit{green}, the set targets in \textit{red}, the trajectory end points in \textit{yellow} and crashes as \textit{magenta} crosses.}
\label{fig:fig:traj-on-map-sim}
\vspace{-6mm}
\end{figure}

In the following, the performance of the navigation policies is analyzed when deployed in unseen environments in simulation.
We constructed two $\SI{10}{\meter} \times \SI{10}{\meter}$ evaluation maps as shown in Figure~\ref{fig:fig:traj-on-map-sim}:
(i) A test \textit{maze} and (ii) an environment with thin walls and \textit{clutter}.
Then, we conducted the following experiment:
100 random start and target positions were sampled for each of the two environments and consistently used for the evaluation of all models.
% In order to make the results consistent, each model was evaluated given the same randomly generated positions per map.
Possible outcomes for each run are a \textit{success}, a \textit{timeout} or a \textit{crash}.
The timeout is triggered, if the target cannot be reached within \SI{5}{\minute}.
This time would allow the robot to travel \SI{60}{\meter} with an average speed of \SI{0.2}{\meter\per\second} and should suffice to reach the target on a \SI{10}{\meter} $\times$ \SI{10}{\meter} map.
Each episode is aborted after a collision.
The resulting trajectories of the evaluation with model c$_{1000}$+CPO on both maps are visualized in Figure~\ref{fig:fig:traj-on-map-sim}.

\begin{figure*}[htbp]
\centering
    \includegraphics[width=\textwidth,trim=10 7 0 0]{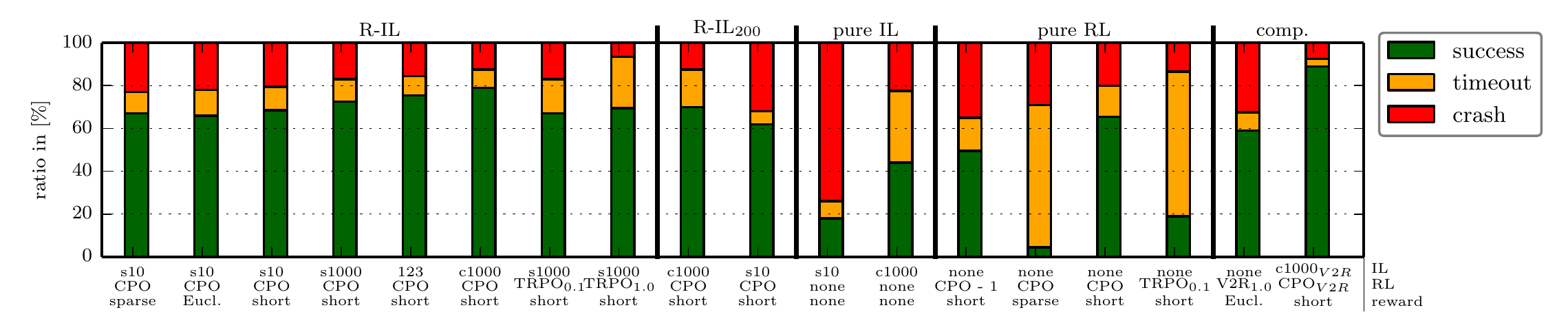}
    \vspace{-5mm}
    \caption{Evaluation results of 200 trajectories on the previously unseen test maps (100 each) as shown in Figure \ref{fig:fig:traj-on-map-sim}.
    The outcome of each trajectory can be a success, a timeout (not reaching the target after \SI{5}{\minute}), or a crash.
    The models are split in five categories: R-IL, where \ac{il} is combined with 1000 \ac{rl} iterations; R-IL$_{200}$;  with 200 \ac{rl} iterations only; pure \ac{il}; pure \ac{rl} and the \textit{comp.} approaches for comparison, comprising the method presented in \cite{tai2017virtual} (V2R) and our method based on the model presented in \cite{tai2017virtual} (c$_{1000}$+CPO$_{V2R}$).
    More details of the analyzed models can be found in Table \ref{tab:models}.
    }
    \label{fig:test_map_eval}
\vspace{-6mm}
\end{figure*}

Based on the 200 evaluation trajectories per model, Figure~\ref{fig:test_map_eval} presents the resulting statistics.
%Compared to Figure~\ref{fig:test_map_eval}, two more models are added under the topic \textit{comp}.
For comparison, first, we trained the model presented in \cite{tai2017virtual} in our environments, which in the following will be referred to as the V2R (virtual-to-real) model.
Second, we used their policy architecture to train our \ac{ril} policy (pretrained in c$_{1000}$) in order to test the generalization to other model structures (c$_{1000}$+CPO$_{V2R}$).
The robot's velocity was removed from the inputs (resulting in 12 inputs) as supervised learning approaches (as for pre-training) tend to predict the prior velocity values instead of focussing on the perception \cite{muller2005off}.

Figure~\ref{fig:test_map_eval} shows that more reward information during training and more pre-training samples not only benefit the training but also the generalization performance.
c$_{1000}$+CPO, the model with shortest distance reward and \textit{complex} pre-training, shows the best generalization performance to unseen environments (using the model structure shown in Figure~\ref{fig:nn-structure}), with a success rate of 79\%.
Interestingly, even the model with only sparse reward and 10 demonstration trajectories in the \textit{simple} environment shows similar performance to the fixed collision penalty TRPO methods, which were pre-trained with 1000 samples and use the full reward.
Both R-IL TRPO methods show a lower success rate than the corresponding CPO model (s$_{1000}$+CPO), which also shows that encoding both collision avoidance and reaching the target in one reward is inferior to encoding the collision avoidance as a constraint.
Furthermore, the R-IL$_{200}$ models show that early stopping of the training (at 200 RL iterations) still leads to similar performance as training pure RL from scratch.
Therefore, pre-training allows for a RL training time reduction of around 80\% in order to achieve the same performance.
CPO1 model, which reached a high success rate during training, does not generalize properly to unseen and more complex environments.

The V2R method \cite{tai2017virtual} (second-to-right bar) shows a similar success rate as the CPO123 model, while the crash rate is about 50\% higher although a collision penalty of 1.0 was used.
However, it uses the Euclidean distance reward which is a slight disadvantage compared to CPO123.
With V2R, the same problems as with other fixed collision penalty methods can be observed, which is the difficult tuning between exploration and collision avoidance.
Our approach also generalizes well to other model structures as the one presented in \cite{tai2017virtual}, as shown by the rightmost bar of Figure~\ref{fig:test_map_eval}.
Using this simpler architecture, the success rate can even be further improved in our test scenarios, which leaves more room for further graph optimization, which is not covered in this paper.
% The \ac{ril} models also show improvements in terms of both success and crash ratio compared to the V2R approach.
%\todomark{Why is ddpg better than pure RL CPO?}

% Overall it can be said, that \ac{ril} with constraint-based training improves the robustness of map-less navigation methods, both during training and testing in unseen environments.
%
% Overall it can be said, that \ac{ril} improves the navigation performance compared to its plain counterparts.
% Plain \ac{il} is target driven but suffers from a high crash rate.
% Plain \ac{rl} ends up with a lower crash rate due to the collision constraint yet struggles to reach the target.
% \ac{ril} combines the best of both worlds by making the exploration more efficient and ends up with a target driven but safer navigation solution.
% By only taking into account local measurements and a relative target, the final \ac{ril} models end up with a success rate of at least 70\% in previously unseen environments.

\subsection{Real-world experiments}
Moving to the real world scenarios further shows the generalization capabilities of the models and also their robustness against sensor noise and actuation delays.
%The platform used for our real-world tests was described in Section~\ref{subsec:experimental-setup}.
The models are purely trained in simulation and the real-world test environment is unknown to the agents.

\begin{figure}[t]
\centering
    \includegraphics[width=0.8\columnwidth,trim=10 15 0 5]{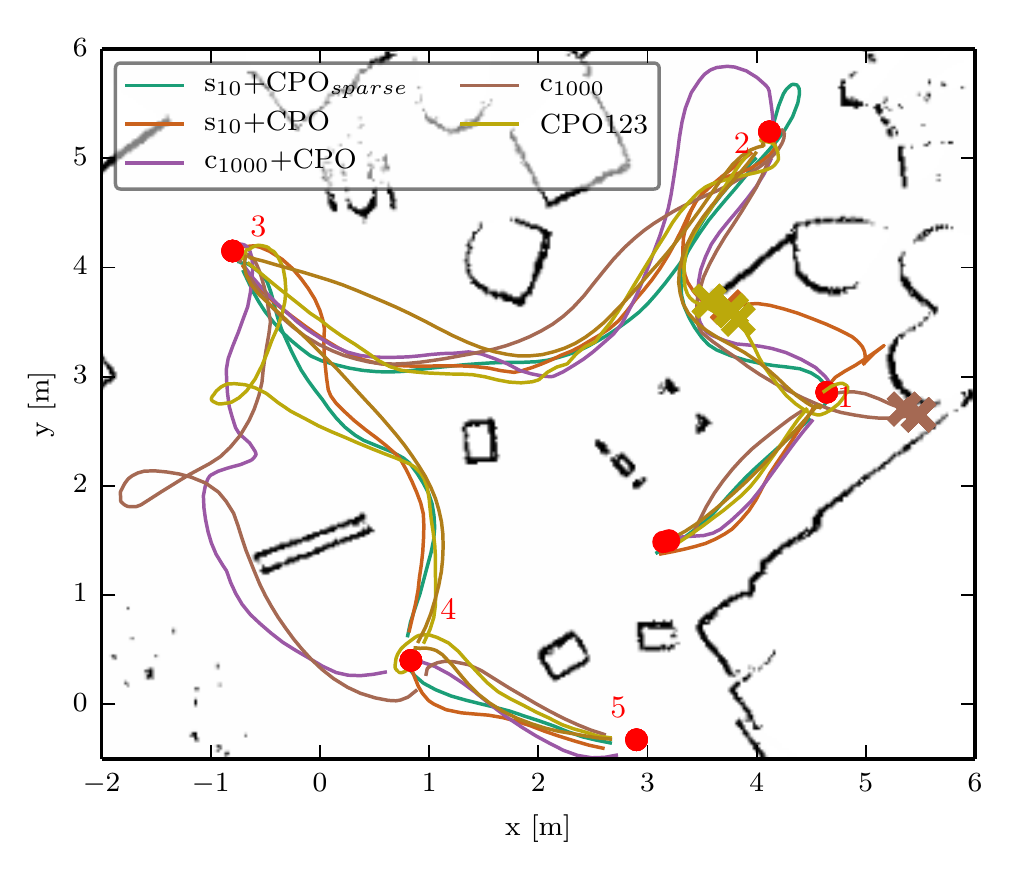}
    \vspace{-1mm}
    \caption{Trajectories driven with the real robotic platform for a subset of the models analyzed in Figure \ref{fig:test_map_eval}.
    \textit{Red dots} depict the numbered target positions, \textit{crosses} in trajectory colors show crashes of the corresponding agents.
    For clarity reasons, only the first out of 5 runs with each model is shown.}
    \label{fig:traj-real}
\vspace{-2mm}
\end{figure}

A quantitative analysis of the trajectories is provided in Table \ref{tab:real-world-results}, where the number of crashes, the amount of manual joystick interference and the comparison of the learning-based trajectories compared to the ones taken by the grid-based \texttt{move\_base} planning module (which uses global map information) are listed.
% For each model we conducted five runs.
Table~\ref{tab:real-world-results} both lists the average and maximum values observed during five runs per model.
The human joystick interference was triggered, if no motion command was sent by the autonomous agent for 10 seconds.

The pure \ac{rl} model tends to be more cautious, which results in a larger factor $\lambda t_{\text{MB}}$, which is the relative time compared to a global planner.
The pure \ac{il} model collides more often as there is no collision constraint or penalty during training.
Also the \ac{ril} models generalize well to the unseen real-world environment and show similar performance.
As expected, c$_{1000}$+CPO shows the best performance.
However, s$_{10}$+CPO$_{sparse}$ performs surprisingly well.
This can be explained by the fact that the sparse reward structure allows for the best generalization performance to unseen environments, since no information about the shortest path to the goal has to be inferred.
This is a promising result, as for this model no environment information and reward shaping is required.
By combining sparse reward with pre-training and constraint-based RL, even real-world training might be feasible.

% Although the amount of crashes is the same, Figure~\ref{fig:traj-real} and our observations during the tests showed that the type of crashes under the \ac{il} policy are significantly worse than under the RL123 policy.
% While the RL123 agent only slightly brushes the walls on its side, the \ac{il} agent fully hits the obstacles.
% This clearly shows the benefit of using \ac{rl}, which can actively constrain collisions during training and does not only resort to an implicit collision avoidance by expert imitation.
%
% The \ac{ril} models show a more robust and better performance than the other models, in comparison with the graph-based planner.
% In addition, the agent never gets stuck and no manual joystick interference was required, both with the RL123 and \ac{ril} agents.
% The \ac{ril} agents are more cautious than the expert planner, which is reflected in the longer trajectory durations (higher $\lambda t_{\text{MB}}$) and results from the enforced collision constraint during training.
%

\begin{table}[btp]
\vspace{2mm}
\caption{Average results (5 runs) from the real-world experiments, as shown in Figure~\ref{fig:traj-real}.
The corresponding maximum values are listed in parenthesis.
d$_{RC}$ stands for the remote controlled (joystick) distance, $\lambda d_{\text{MB}}$ for the relative distance compared to \texttt{move\_base} and $\lambda t_{\text{MB}}$ for the relative time compared to \texttt{move\_base}}
\label{tab:real-world-results}
\vspace{-3mm}
\begin{center}
\begin{tabular}{|c||c|c|c|c|  }
 \hline
 model   & \#crash & d$_{\text{RC}}$ [m] & $\lambda d_{\text{MB}}$  & $\lambda t_{\text{MB}}$\\
 \hline
 % shapes10+rl
 %   Distance: 20.817
 %   Time: 171.130
 %   Number of crashes: 0.800
 %   Distance joystick: 0.151
 %   Relative distance joystick: 0.007
 %   Relative distance ROS: 1.172
 %   Relative time ROS: 1.864
 s$_{10}$+CPO          & 0.8 \tiny{(2)}  & 0.15 \tiny{(0.28)} & 1.17 \tiny{(1.2)}   & 1.86 \tiny{(1.95)} \\
 % office1000+rl
 %   Distance: 21.074
 %   Time: 95.330
 %   Number of crashes: 0.000
 %   Distance joystick: 0.000
 %   Relative distance joystick: 0.000
 %   Relative distance ROS: 1.186
 %   Relative time ROS: 1.038
 c$_{1000}$+CPO        & 0.0 \tiny{(0)}  & 0.0 \tiny{(0)}  & 1.19 \tiny{(1.22)} & 1.04 \tiny{(1.24)} \\
 % shapes_10_sparse_reward
 %   Distance: 20.381
 %   Time: 127.050
 %   Number of crashes: 0.000
 %   Distance joystick: 0.007
 %   Relative distance joystick: 0.000
 %   Relative distance ROS: 1.147
 %   Relative time ROS: 1.384
 s$_{10}$+CPO$_{sparse}$ & 0.0 \tiny{(0)}  & 0.01 \tiny{(0.03)} & 1.15 \tiny{(1.19)}   & 1.38  \tiny{(2.00)} \\
 % il
 %   Distance: 22.938
 %   Time: 160.970
 %   Number of crashes: 1.600
 %   Distance joystick: 0.051
 %   Relative distance joystick: 0.002
 %   Relative distance ROS: 1.291
 %   Relative time ROS: 1.753
 c$_{1000}$           & 1.6 \tiny{(4)}  & 0.05 \tiny{(0.12)} & 1.29  \tiny{(1.39)} & 1.75 \tiny{(2.52)} \\
 % rl
 %   Distance: 22.327
 %   Time: 195.290
 %   Number of crashes: 0.600
 %   Distance joystick: 0.076
 %   Relative distance joystick: 0.003
 %   Relative distance ROS: 1.257
 %   Relative time ROS: 2.127
 CPO123                & 0.6 \tiny{(2.0)}  & 0.08 \tiny{(0.15)}  & 1.26 \tiny{(1.29)}  & 2.13 \tiny{(2.18)} \\
 \hline
\end{tabular}
\end{center}
\vspace{-8mm}
\end{table}

%%%%%%%%%%%%%%%%%%%%%%%%% Conclusion %%%%%%%%%%%%%%%%%%%%%%%%%%%%

\section{Conclusion}
\label{sec:conclusion}

In this work, we presented a case study for a learning-based approach for map-less target driven navigation.
It is based on an end-to-end neural network model which maps from raw sensor measurements and a relative target location to motion commands of a robotic platform and is trained using a combination of imitation (IL) and reinforcement learning (RL).
We compare different combinations of prior demonstrations for IL, different RL algorithms and analyze the influence of different reward structures.

Our simulation and real-world experiments show that target-driven demonstrations through IL significantly improve the exploration during RL.
The RL training time in \ac{ril} can be reduced by around 80\% while still achieving similar final performance in terms of success rate and collision avoidance.
While pure RL does achieve the same collision avoidance capabilities as \ac{ril}, there are significant differences in the target reaching success.
Pre-training with supervised IL provides a good intuition for more efficient exploration during RL, even if only 10 demonstrations are provided.
This becomes even more pronounced when using low information reward structures, like sparse target reward.

Furthermore, our experiments show that constraint-based methods focus on enforcing the collision constraint early during training.
This makes exploration harder yet allows for safer training and deployment which becomes important when moving towards real-world applications.
Therefore, especially in combination with IL, to achieve safe navigation capabilities, we recommend to enforce collision avoidance by constraint instead of a fixed penalty in the reward signal.

Our trained navigation models are able to reliably navigate in unseen environments, both in simulation and the real world.
We do not recommend to replace global planning if a map is available, yet this work shows the current state of what is possible using only local information for navigation scenarios, where no environment map is available.

While in this work, training was purely conducted in simulation, in future work we will investigate how real-world human demonstrations can be leveraged and how this navigation method can be extended to dynamic environments.

\footnotesize
\bibliographystyle{style/IEEEtran}
\balance
\bibliography{bib/IEEEfull.bib,bib/reinforcement_navigation.bib}

\begin{thebibliography}{10}
\providecommand{\url}[1]{#1}
\csname url@rmstyle\endcsname
\providecommand{\newblock}{\relax}
\providecommand{\bibinfo}[2]{#2}
\providecommand\BIBentrySTDinterwordspacing{\spaceskip=0pt\relax}
\providecommand\BIBentryALTinterwordstretchfactor{4}
\providecommand\BIBentryALTinterwordspacing{\spaceskip=\fontdimen2\font plus
\BIBentryALTinterwordstretchfactor\fontdimen3\font minus
  \fontdimen4\font\relax}
\providecommand\BIBforeignlanguage[2]{{%
\expandafter\ifx\csname l@#1\endcsname\relax
\typeout{** WARNING: IEEEtran.bst: No hyphenation pattern has been}%
\typeout{** loaded for the language `#1'. Using the pattern for}%
\typeout{** the default language instead.}%
\else
\language=\csname l@#1\endcsname
\fi
#2}}

\bibitem{lavalle2006planning}
S.~M. LaValle, \emph{Planning algorithms}.\hskip 1em plus 0.5em minus
  0.4em\relax Cambridge university press, 2006.

\bibitem{pfeiffer2017perception}
M.~Pfeiffer, M.~Schaeuble, J.~Nieto, R.~Siegwart, and C.~Cadena, ``From
  perception to decision: A data-driven approach to end-to-end motion planning
  for autonomous ground robots,'' in \emph{IEEE Int. Conf. on Robotics and
  Automation (ICRA)}.\hskip 1em plus 0.5em minus 0.4em\relax IEEE, 2017, pp.
  1527--1533.

\bibitem{tai2017virtual}
L.~Tai, G.~Paolo, and M.~Liu, ``Virtual-to-real deep reinforcement learning:
  Continuous control of mobile robots for mapless navigation,'' in
  \emph{IEEE/RSJ Int. Conf. on Intelligent Robots and Sys. (IROS)}.\hskip 1em
  plus 0.5em minus 0.4em\relax IEEE, 2017, pp. 31--36.

\bibitem{muller2005off}
U.~Muller, J.~Ben, E.~Cosatto, B.~Flepp, and Y.~L. Cun, ``Off-road obstacle
  avoidance through end-to-end learning,'' in \emph{Advances in neural
  information processing systems}, 2005, pp. 739--746.

\bibitem{mnih2016asynchronous}
V.~Mnih, A.~P. Badia, M.~Mirza, A.~Graves, T.~Lillicrap, T.~Harley, D.~Silver,
  and K.~Kavukcuoglu, ``Asynchronous methods for deep reinforcement learning,''
  in \emph{Int. Conf. on Machine Learning}, 2016, pp. 1928--1937.

\bibitem{kuefler2017imitating}
A.~Kuefler, J.~Morton, T.~Wheeler, and M.~Kochenderfer, ``Imitating driver
  behavior with generative adversarial networks,'' in \emph{Prof. of the
  Intelligent Vehicles Symposium (IV)}.\hskip 1em plus 0.5em minus 0.4em\relax
  IEEE, 2017, pp. 204--211.

\bibitem{grimes2009learning}
D.~B. Grimes and R.~P. Rao, ``Learning actions through imitation and
  exploration: Towards humanoid robots that learn from humans,'' in
  \emph{Creating Brain-Like Intelligence}.\hskip 1em plus 0.5em minus
  0.4em\relax Springer, 2009, pp. 103--138.

\bibitem{achiam2017cpo}
J.~Achiam, D.~Held, A.~Tamar, and P.~Abbeel, ``Constrained policy
  optimization,'' \emph{arXiv preprint arXiv:1705.10528}, 2017.

\bibitem{Abbeel2008}
P.~Abbeel, D.~Dolgov, A.~Ng, and S.~Thrun, ``{Apprenticeship learning for
  motion planning with application to parking lot navigation},'' in
  \emph{{IEEE/RSJ Int. Conf. on Intelligent Robots and Sys. (IROS)}}, Nice,
  France, Sept. 2008, pp. 1083--1090.

\bibitem{pfeiffer2016iros}
M.~Pfeiffer, U.~Schwesinger, H.~Sommer, E.~Galceran, and R.~Siegwart,
  ``Predicting actions to act predictably: Cooperative partial motion planning
  with maximum entropy models,'' in \emph{{IEEE/RSJ Int. Conf. on Intelligent
  Robots and Sys. (IROS)}}.\hskip 1em plus 0.5em minus 0.4em\relax IEEE, Oct.
  2016, pp. 2096--2101.

\bibitem{kretzschmar2016social}
H.~Kretzschmar, M.~Spies, C.~Sprunk, and W.~Burgard, ``Socially compliant
  mobile robot navigation via inverse reinforcement learning,'' \emph{The Int.
  Journal of Robotics Research}, vol.~35, no.~11, pp. 1289--1307, 2016.

\bibitem{wulfmeier2016watch}
M.~Wulfmeier, D.~Z. Wang, and I.~Posner, ``Watch this: Scalable cost-function
  learning for path planning in urban environments,'' in \emph{Proc of IEEE/RSJ
  Int. Conf. on Intelligent Robots and Sys. (IROS)}.\hskip 1em plus 0.5em minus
  0.4em\relax IEEE, 2016, pp. 2089--2095.

\bibitem{chen2015deepdriving}
C.~Chen, A.~Seff, A.~Kornhauser, and J.~Xiao, ``Deepdriving: Learning
  affordance for direct perception in autonomous driving,'' in \emph{IEEE Int.
  Conf. on Computer Vision (ICCV)}, 2015, pp. 2722--2730.

\bibitem{kim2015deep}
D.~K. Kim and T.~Chen, ``Deep neural network for real-time autonomous indoor
  navigation,'' \emph{arXiv preprint arXiv:1511.04668}, 2015.

\bibitem{sergeant2015autoencoders}
J.~Sergeant, N.~S{\"u}nderhauf, M.~Milford, and B.~Upcroft, ``Multimodal deep
  autoencoders for control of a mobile robot,'' in \emph{Proc. of Australasian
  Conf. for Robotics and Automation (ACRA)}, 2015.

\bibitem{ross2011dagger}
S.~Ross, G.~Gordon, and D.~Bagnell, ``A reduction of imitation learning and
  structured prediction to no-regret online learning,'' in \emph{fourteenth
  Int. Conf. on artificial intelligence and statistics}, 2011, pp. 627--635.

\bibitem{ross2013learning}
S.~Ross, N.~Melik-Barkhudarov, K.~S. Shankar, A.~Wendel, D.~Dey, J.~A. Bagnell,
  and M.~Hebert, ``Learning monocular reactive uav control in cluttered natural
  environments,'' in \emph{IEEE Int. Conf. on Robotics and Automation (ICRA),
  2013}.\hskip 1em plus 0.5em minus 0.4em\relax IEEE, 2013, pp. 1765--1772.

\bibitem{ho2016generative}
J.~Ho and S.~Ermon, ``Generative adversarial imitation learning,'' in
  \emph{Advances in Neural Inform. Processing Sys.}, 2016, pp. 4565--4573.

\bibitem{tai2017socially}
L.~Tai, J.~Zhang, M.~Liu, and W.~Burgard, ``Socially-compliant navigation
  through raw depth inputs with generative adversarial imitation learning,''
  \emph{arXiv preprint arXiv:1710.02543}, 2017.

\bibitem{Bischoff2013HierarchicalRL}
B.~Bischoff, D.~Nguyen-Tuong, I.-H. Lee, F.~Streichert, and A.~Knoll,
  ``Hierarchical reinforcement learning for robot navigation,'' in
  \emph{ESANN}, 2013.

\bibitem{Zuo2014ARL}
B.~Zuo, J.~Chen, L.~Wang, and Y.~Wang, ``A reinforcement learning based robotic
  navigation system,'' \emph{IEEE Int. Conf. on Sys., Man, and Cybernetics
  (SMC)}, pp. 3452--3457, 2014.

\bibitem{mirowski2016learning}
P.~Mirowski, R.~Pascanu, F.~Viola, H.~Soyer, A.~J. Ballard, A.~Banino,
  M.~Denil, R.~Goroshin, L.~Sifre, K.~Kavukcuoglu, \emph{et~al.}, ``Learning to
  navigate in complex environments,'' \emph{Prof. of the Int. Conf. on Learning
  Representations}, 2017.

\bibitem{Bruce2017OneShotRL}
J.~Bruce, N.~S{\"u}nderhauf, P.~W. Mirowski, R.~Hadsell, and M.~Milford,
  ``One-shot reinforcement learning for robot navigation with interactive
  replay,'' \emph{CoRR}, vol. abs/1711.10137, 2017.

\bibitem{Zhang2017DeepRL}
J.~Zhang, J.~T. Springenberg, J.~Boedecker, and W.~Burgard, ``Deep
  reinforcement learning with successor features for navigation across similar
  environments,'' \emph{IEEE/RSJ Int. Conf. on Intelligent Robots and Sys.
  (IROS)}, pp. 2371--2378, 2017.

\bibitem{zhu2017target}
Y.~Zhu, R.~Mottaghi, E.~Kolve, J.~J. Lim, A.~Gupta, L.~Fei-Fei, and A.~Farhadi,
  ``Target-driven visual navigation in indoor scenes using deep reinforcement
  learning,'' in \emph{IEEE Int. Conf. on Robotics and Automation
  (ICRA)}.\hskip 1em plus 0.5em minus 0.4em\relax IEEE, 2017, pp. 3357--3364.

\bibitem{chen2017socialdrl}
Y.~F. Chen, M.~Everett, M.~Liu, and J.~P. How, ``Socially aware motion planning
  with deep reinforcement learning,'' \emph{CoRR}, vol. abs/1703.08862, 2017.

\bibitem{Balaguer2011CombiningIA}
B.~Balaguer and S.~Carpin, ``Combining imitation and reinforcement learning to
  fold deformable planar objects,'' \emph{IEEE/RSJ Int. Conf. on Intelligent
  Robots and Sys. (IROS)}, pp. 1405--1412, 2011.

\bibitem{Zhu2018ReinforcementAI}
Y.~Zhu, Z.~Wang, J.~Merel, A.~A. Rusu, T.~Erez, S.~Cabi, S.~Tunyasuvunakool,
  J.~Kram{\'a}r, R.~Hadsell, N.~de~Freitas, and N.~Heess, ``Reinforcement and
  imitation learning for diverse visuomotor skills,'' \emph{CoRR}, vol.
  abs/1802.09564, 2018.

\bibitem{williams1992simple}
R.~J. Williams, ``Simple statistical gradient-following algorithms for
  connectionist reinforcement learning,'' in \emph{Reinforcement
  Learning}.\hskip 1em plus 0.5em minus 0.4em\relax Springer, 1992, pp. 5--32.

\bibitem{schulman2015trust}
J.~Schulman, S.~Levine, P.~Abbeel, M.~Jordan, and P.~Moritz, ``Trust region
  policy optimization,'' in \emph{Int. Conf. on Machine Learning}, 2015, pp.
  1889--1897.

\bibitem{Berkenkamp2017SafeMR}
F.~Berkenkamp, M.~Turchetta, A.~Schoellig, and A.~Krause, ``Safe model-based
  reinforcement learning with stability guarantees,'' in \emph{Advances in
  Neural Inform. Process. Sys.}, 2017, pp. 908--918.

\bibitem{ros}
M.~Quigley, K.~Conley, B.~Gerkey, J.~Faust, T.~Foote, J.~Leibs, R.~Wheeler, and
  A.~Y. Ng, ``Ros: an open-source robot operating system,'' in \emph{ICRA
  workshop on open source software}.\hskip 1em plus 0.5em minus 0.4em\relax
  Kobe, Japan, 2009, p.~5.

\bibitem{vaughan2008massively}
R.~Vaughan, ``Massively multi-robot simulation in stage,'' \emph{Swarm
  intelligence}, vol.~2, no. 2-4, pp. 189--208, 2008.

\end{thebibliography}

\end{document}